\documentclass[11pt,a4paper]{article}
\usepackage[hyperref]{emnlp2020}
\usepackage{times}
\usepackage{latexsym}

\usepackage{graphicx}
\usepackage{eurosym}
\usepackage{tablefootnote}
\usepackage{kantlipsum}
\usepackage{tabularx}
\graphicspath{ {images/} }
\usepackage{hyperref}
\usepackage{amsmath}
\usepackage{hhline}
\usepackage[utf8]{inputenc}
\usepackage{multirow}

\usepackage{algorithm}
\usepackage{algorithmicx}
\usepackage[noend]{algpseudocode}
\usepackage{multicol}

\usepackage[T1]{fontenc}
\usepackage{listings}
\usepackage{stfloats}
\usepackage{csquotes}
\usepackage{array}
\usepackage{booktabs}
\usepackage[super]{nth}
\usepackage[inline,shortlabels]{enumitem}
\usepackage{enumitem}
\usepackage{makecell}
\usepackage{siunitx}
\usepackage{xcolor, colortbl}

\usepackage{svg}
\usepackage{float}
\usepackage{ amssymb }

\usepackage[hang,flushmargin]{footmisc}

\usepackage{microtype}

\aclfinalcopy 

\title{Domain Adversarial Fine-Tuning as an Effective Regularizer}

\author{ 
	Giorgos Vernikos$^{1}$, 
	Katerina Margatina$^{2}$,
	Alexandra Chronopoulou$^{3}$,
	Ion Androutsopoulos$^{4}$\\\\
	$^1$ School of ECE, National Technical University of Athens, Greece \\
	$^2$ Computer Science Department, University of Sheffield, UK\\
	$^3$ Center for Information and Language Processing, LMU Munich, Germany \\
	$^4$ Department of Informatics, Athens University of Economics and Business, Greece \\  
    {\tt}\\
	{\tt gvernikos@mail.ntua.gr, k.margatina@sheffield.ac.uk, achron@cis.lmu.de} \\
	{\tt  ion@aueb.gr}
	}

\begin{document}
\maketitle
\begin{abstract}
In Natural Language Processing (NLP), pretrained language models (LMs) that are transferred to downstream tasks have been recently shown to 
achieve state-of-the-art results. However, standard fine-tuning can 
degrade the general-domain representations captured during pretraining. To address this issue,
we introduce a new regularization technique, \textsc{after}; domain Adversarial Fine-Tuning as an Effective Regularizer. 
Specifically, we complement the task-specific loss used during fine-tuning with an adversarial objective. 
This additional
loss term is related to an adversarial classifier, that aims to discriminate between \emph{in-domain} and \emph{out-of-domain} text representations.  
In-domain refers to the \textit{labeled} dataset of the task at hand while out-of-domain refers to \textit{unlabeled} data from a different domain.
Intuitively, the adversarial classifier 
acts as a regularizer which prevents the model from overfitting to the task-specific domain.
Empirical results on 
various natural language understanding tasks
show that \textsc{after} leads to improved performance compared to standard fine-tuning.

\end{abstract}

\section{Introduction}
Current research in NLP focuses on transferring knowledge from a 
language model (LM), pretrained on large general-domain data, to a 
target 
task. The LM representations are transferred to the target task
either 
as additional features of a task-specific model ~\cite{peters-etal-2018-deep}, or
by fine-tuning ~\cite{howard-ruder-2018-universal,  devlin-etal-2019-bert, NIPS2019_8812}.
Standard fine-tuning involves initializing the target model with the pretrained LM and training it with the target data.

Fine-tuning, however, can lead to catastrophic forgetting \cite{Goodfellow2013AnEI}, if the pretrained 
LM
representations are adjusted to such an extent to the target task,
that most generic knowledge, captured 
during pretraining, is in effect forgotten \cite{howard-ruder-2018-universal}. 
A related problem of fine-tuning
is overfitting to the target task, that often occurs when
only a small number of training examples is available~\cite{NIPS2015_5949}.

Adversarial training is a method to increase robustness and 
regularize deep neural networks \cite{Goodfellow2014ExplainingAH, Miyato2017-rs}. 
It has been used for domain adaptation \cite{Ganin:2016:DTN:2946645.2946704} to train a model from scratch to produce representations that are invariant to different domains. Inspired by this approach,
we propose a regularization technique for the fine-tuning process of a pretrained LM,
that aims to optimize knowledge transfer
to the target task and avoid overfitting. 

Our method, domain Adversarial Fine-Tuning as an Effective Regularizer (\textsc{after}) extends standard fine-tuning by adding an 
adversarial objective to the task-specific loss. We leverage out-of-domain \emph{unlabeled} data (i.e. from a different domain than the target task domain).  The transferred LM is fine-tuned so that an adversarial classifier cannot discriminate between
text representations from 
in-domain and out-of-domain data. This loss aims to regularize the extent to which the model representations are allowed to adapt to the target task domain. Thus, \textsc{after} is able to  preserve the general-domain knowledge acquired during the pretraining of the LM, while adapting to the target task.

Our contributions are:
(1) We propose \textsc{after}, an LM 
fine-tuning 
method that aims to avoid catastrophing forgetting of general-domain knowledge, acting as a new kind of regularizer. 
(2) We show that \textsc{after} improves the performance of standard fine-tuning in four  natural language understanding tasks from the GLUE benchmark~\cite{wang-etal-2018-glue}, with two different pretrained LMs: \textsc{BERT}~\cite{devlin-etal-2019-bert}, and \textsc{XLNet}~\cite{NIPS2019_8812}.
(3) We further conduct an ablation study
to provide useful insights regarding the key factors of the proposed approach.
\section{Related Work}
Several approaches have been proposed for the adaptation of a model trained on a 
domain $\mathcal{D}_S$ to a
different domain $\mathcal{D}_T$, where no labeled data is available \cite{Grauman:2012:GFK:2354409.2355024,DBLP:journals/corr/TzengHZSD14, Sun:2016:RFE:3016100.3016186}. 
\citet{Ganin:2016:DTN:2946645.2946704}
were the first to propose adversarial training 
for \textit{domain adaptation}. 
They introduced a gradient reversal layer to
adversarially 
train a classifier that should not be able to discriminate between $\mathcal{D}_S$ and
$\mathcal{D}_T$,
in image classification and sentiment analysis tasks.

Various 
 adversarial losses have been used for
domain adaptation in several NLP tasks, such as question answering \cite{lee2019domainagnostic}, machine reading comprehension \cite{wang2019adversarial} and 
cross-lingual named entity recognition \cite{keung-etal-2019-adversarial}.
Adversarial approaches have been also used to learn latent representations that are agnostic to different attributes of the input text, such as language ~\cite{DBLP:conf/iclr/LampleCDR18, DBLP:conf/iclr/LampleCRDJ18} and style ~\cite{UTTS}.
Contrary to previous \emph{domain adaptation} work,
we 
explore the addition of an adversarial loss term
to serve 
as a
\textit{regularizer} for fine-tuning. 

Other variants of LM fine-tuning include a supplementary supervised training stage in data-rich tasks ~\cite{Phang2018SentenceEO} or multi-task learning with additional supervised tasks ~\cite{liu-etal-2019-multi-task}. However, such methods require additional \textit{labeled} data.
A common way to leverage \textit{unlabeled} data
during fine-tuning is 
through an additional stage of language modeling. 
For this stage, the unlabeled data can either come from the
task-specific dataset (i.e. the labels are dropped and language modelling is performed on the input data)~\cite{howard-ruder-2018-universal},
or additional unlabeled in-domain corpora 
~\cite{Sun2019HowTF, Gururangan2020DontSP}.
This approach adds a computationally expensive step that requires unlabeled data from a specific source. By contrast, our method leverages out-of-domain data with only  a small computational overhead and minimal changes to the fine-tuning process. 

Our work is compatible with the semi-supervised learning paradigm \cite{SSL} that combines learning from both labeled and unlabeled data. In this setting, unlabeled data from the task domain is leveraged using a consistency loss which enforces invariance of the output given small perturbations of the input ~\cite{Miyato2017-rs, clark-etal-2018-semi}. The adversarial loss term of \textsc{after} can be interpreted as a consistency loss that ensures invariance of representations across domains. 

Recently, adversarial or trust region based approaches \cite{Zhu2020FreeLB:, jiang-etal-2020-smart, DBLP:journals/corr/abs-2008-03156} have been proposed as an extension to the LM fine-tuning process. These methods introduce constraints that prevent aggressive updating of the pretrained parameters or enforce smoothness during fine-tuning. However, these approaches require additional forward and backward computations while our method is more computationally efficient and can be implemented with minimal changes to the fine-tuning procedure.

\section{Proposed Approach}
\begin{figure*}[t]
        \centering
         \includegraphics[width=0.8\textwidth]
         {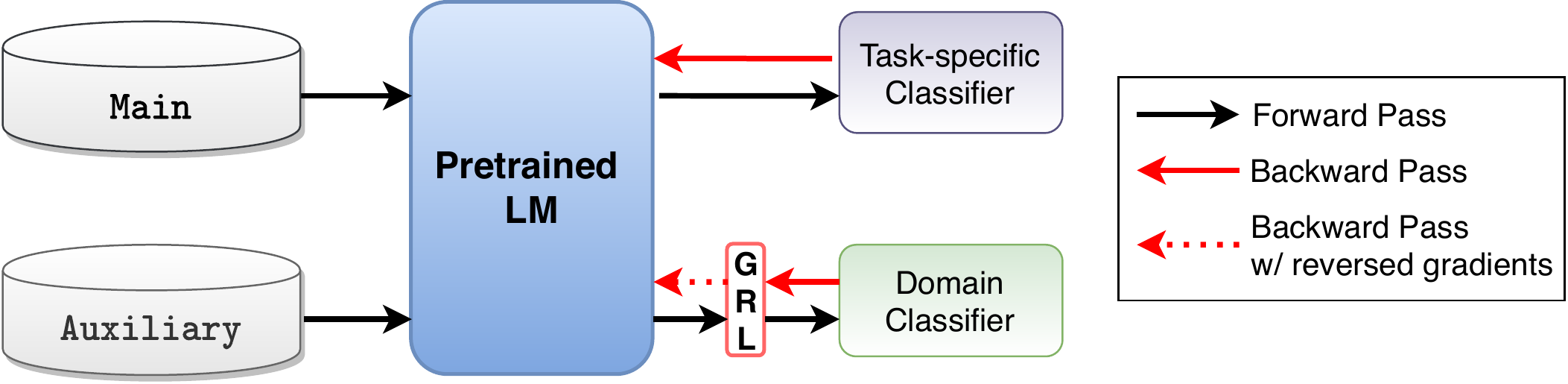}
        \caption[]{Illustration of the proposed approach,  \textsc{after}. The task-specific classifier leverages the labeled data from the downstream task (\texttt{Main}) while the domain classifier uses  unlabeled
        data from both \texttt{Main} and \texttt{Auxiliary} datasets as well as the created domain labels.
        }
        \label{fig:AFTER_fig}
\end{figure*}

Fig.~\ref{fig:AFTER_fig} provides a 
high-level 
overview of \textsc{after}.

\vspace{2pt}
\noindent\textbf{Problem Definition}.
We tackle a \texttt{Main} task, with a labeled dataset from domain $\mathcal{D_\textit{M}}$.
We further exploit an existing \emph{unlabeled corpus}, \texttt{Auxiliary}, that comes from a different domain $\mathcal{D}_{\textit{AUX}}$.
We label each sample 
with the corresponding domain label 
$y_{\mathcal{D}}$, 
$y_{\mathcal{D}}=0$ for samples from \texttt{Main}, and $y_{\mathcal{D}}=1$
for samples from \texttt{Auxiliary}.  
We 
note that we 
\textit{do not} use any real labels from \texttt{Auxiliary} 
(if there are any).
The domain labels are used to train a classifier that 
discriminates between $\mathcal{D_\textit{M}}$ and $\mathcal{D}_{\textit{AUX}}$.

\vspace{2pt}
\noindent\textbf{Model}.
We initialize our model with pretrained weights from a top-performing language model, such as \textsc{BERT}~\cite{devlin-etal-2019-bert} or \textsc{XLNet}~\cite{NIPS2019_8812}.
The representation of both \textsc{BERT} and \textsc{XLNet} for the input sequence  is encoded in the \texttt{[CLS]} token output embedding.
We add a linear
layer on top of the
sequence representation
(\texttt{[CLS]} output embedding)
for the \texttt{Main} task, resulting in a task-specific loss $L_{Main}$.
We also add another linear layer for the 
binary domain classifier (Figure~\ref{fig:AFTER_fig}), 
with a corresponding loss $L_{Domain}$,
which has the same input.

\vspace{2pt}
\noindent\textbf{Adversarial Fine-tuning}.
The domain discriminator outputs a 
domain label
for each sample of the training set.
We seek representations that are both discriminative for the \texttt{Main} task and indiscriminative for the domain classifier. Hence, we minimize $L_{Main}$ and at the same time maximize $L_{Domain}$, by fine-tuning the pretrained LM with the joint loss:
\begin{equation}
    \mathcal{L}_\text{AFTER}=L_{Main} - \lambda L_{Domain}
    \label{eq:loss}
\end{equation}
where $\lambda$ ($\lambda > 0$) controls
the importance of the domain loss.
The parameters of the domain classifier are trained to predict the (true) domain label, while the rest of the network is trained to mislead the domain classifier, thereby developing domain-independent internal representations.

\vspace{2pt}
\noindent\textbf{Gradient Reversal Layer}. We use a Gradient Reversal Layer (\texttt{GRL}) \cite{Ganin:2016:DTN:2946645.2946704} between
the \texttt{[CLS]} output embedding
and the domain discriminator layer, as shown in Figure~\ref{fig:AFTER_fig}, to maximize $L_{Domain}$.
During the forward pass, \texttt{GRL} acts as an identity transform, but during backpropagation, \texttt{GRL} \textit{reverses} the 
gradients. 
In effect, the pretrained LM parameters are updated towards the opposite direction of the gradient of $L_{Main}$ and, adversarially, towards the direction of the gradient of $L_{Domain}$.

\section{Experiments}\label{sec:exps}
\noindent\textbf{Datasets}.
We experiment with four \texttt{Main} datasets  from the GLUE benchmark ~\cite{wang-etal-2018-glue}.
The chosen datasets represent the broad variety of natural language understanding tasks,
such as linguistic acceptability (\textsc{CoLA}) \cite{warstadt2018neural}, sentiment analysis (\textsc{SST-2}) \cite{socher-EtAl:2013:EMNLP}, paraphrase detection (\textsc{MRPC}) \cite{dolan-brockett-2005-automatically} and textual entailment (\textsc{RTE}) \cite{RTE, RTE2, RTE3, RTE5}. The datasets used represent both  high (\textsc{SST-2}) and low-resource (\textsc{RTE}, \textsc{CoLA}, \textsc{MRPC}) tasks, as well as single-sentence (\textsc{CoLA}, \textsc{SST-2}) and sentence-pair (\textsc{MRPC}, \textsc{RTE}) tasks. 
For \texttt{Auxiliary} data we select corpora from various domains. For 
the \textsc{news} domain we use the \textsc{ag news} dataset \cite{Zhang:2015:CCN:2969239.2969312} and for 
the 
\textsc{reviews} domain we use a part of the Electronics reviews of  \citet{He:2016:UDM:2872427.2883037}. For 
the 
\textsc{legal} domain we use 
the English part of \textsc{Europarl}~\cite{article} and for
the 
\textsc{medical} domain we use papers from PubMed, provided by \citet{cohan-etal-2018-discourse}. We also use  math questions from the dataset of \citet{Saxton2019AnalysingMR} for the \textsc{math} domain. Table~\ref{table:datasets} summarizes all datasets.
More details regarding the selection and processing of the datasets can be found in Appendix~\ref{sec:dataset_details}.

\setlength{\tabcolsep}{6pt} 
\renewcommand{\arraystretch}{1.0} 

\begin{table}[t]
\resizebox{\columnwidth}{!}{%

\centering
\small
\begin{tabular}{lcc}
\Xhline{2\arrayrulewidth}
\textsc{Dataset} & \textsc{Domain}  &$N_{train}$  \\ \hline
\multicolumn{3}{l}{\texttt{Main}} \\ \hline
\textsc{CoLA}   & Miscellaneous   &  8.5K\\
\textsc{SST-2}     & Movie Reviews    & 67K\\
\textsc{MRPC}        & News    & 3.7K\\
\textsc{RTE}      & News, Wikipedia   & 2.5K\\ \hline
\multicolumn{3}{l}{\texttt{Auxiliary}} \\ \hline
\textsc{ag news}  & Agricultural News (\textsc{news}) & 120K\\ 
\textsc{europarl}  & Legal Documents (\textsc{legal}) & 120K\\
\textsc{amazon}  & Electronics Reviews (\textsc{reviews})  & 120K\\
\textsc{pubmed}  & Medical Papers (\textsc{medical}) & 120K\\
\textsc{mathematics}  & Mathematics  Questions (\textsc{math}) & 120K\\
\hline
\Xhline{2\arrayrulewidth}
\end{tabular}
 }

\caption{
Datasets used. 
$N_{train}$ denotes the number of training examples. 
The indicator (\textsc{domain}) summarizes the domain of each \texttt{Auxiliary} dataset.
}
\label{table:datasets}
\end{table}

\vspace{2pt}
\noindent\textbf{Baselines}.
We compare our approach (\textsc{after}) with the standard fine-tuning (\textsc{sft}) scheme of the pretrained LMs. 
As our baselines we fine-tune two pretrained LMs (\textsc{BERT-base} and \textsc{XLNet-base}), using the suggested hyperparameters from \citet{devlin-etal-2019-bert} and \citet{NIPS2019_8812} respectively. 

\vspace{2pt}
\noindent\textbf{Implementation Details}.
We base our implementation on Hugging Face's Transformers library ~\cite{wolf2019huggingfaces} in PyTorch \cite{NEURIPS2019_9015}. We make our code publicly available\footnote{\url{https://github.com/GeorgeVern/AFTERV1.0}}. 
We tune the $\lambda$ hyperparameter  of Eq.~\ref{eq:loss} on
the validation set for each experiment, finding that most values of $\lambda$ improve over the baseline.
We fine-tune each model for $4$ epochs and 
evaluate the model $5$ times per epoch, as suggested by \citet{Dodge2020FineTuningPL}. We select the best model based on the validation loss. 
For more implementations details see Appendix~\ref{sec:hyperparameters}.

\begingroup
\setlength{\tabcolsep}{8pt} 
\renewcommand{\arraystretch}{1.02} 

\begin{table*}[t]

\centering
\small
\begin{tabular}{lcccc}
\Xhline{2\arrayrulewidth}

\multirow{2}{*}{} &\textbf{CoLA} &\textbf{SST-2} &\textbf{MRPC} &\textbf{RTE} \\ 

&\textit{Matthews corr.} &\textit{Accuracy} &\textit{Accuracy / F1} &\textit{Accuracy}  \\ \hline

\textsc{BERT sft} &$55.5 \pm 3.2$ &$92.0 \pm 0.5$ &$85.4 \pm 1.1$ / $89.6 \pm 0.6$ &$64.3 \pm 3.1$ \\ \hline 

\textsc{after w/ News} &$\mathbf{57.3} \pm 1.5$ &$\underline{92.5} \pm 0.4$ &$\mathbf{87.5} \pm 1.7$ / $\mathbf{91.1} \pm 1.2$  &$\underline{64.7} \pm 1.9$ \\ 

\textsc{after w/ Reviews} &$\underline{57.1} \pm 1.2$ &$\underline{92.4} \pm 0.3$ &$\underline{86.4} \pm 0.3$ / $\underline{90.1} \pm 0.4$ &$\underline{64.6} \pm 0.8$ \\ 

\textsc{after w/ Legal} &$55.0 \pm 1.5$ &$\underline{92.4} \pm 0.3$ &$\underline{86.6} \pm 0.6$ / $\underline{90.3} \pm 0.5$  &$\mathbf{64.8} \pm 1.9$ \\ 

\textsc{after w/ Medical} &$\underline{55.9} \pm 2.9$ &$\mathbf{92.6} \pm 0.3$ &$\underline{86.9} \pm 1.3$ / $\underline{90.7} \pm 1.0$ &$62.6 \pm 3.4$ \\ 

\textsc{after w/ Math} &$\underline{56.1} \pm 2.8$  &$\underline{92.3} \pm 0.8$ &$\underline{87.3} \pm 0.9$ / $\underline{90.8} \pm 0.7$ &$62.5 \pm 1.3$ \\ \Xhline{2\arrayrulewidth}

\textsc{XLNet sft} & $-$  & $93.0 \pm 0.7$ &$86.4 \pm 0.7$ / $90.1\pm 0.5$ & $64.7 \pm 4.4$ \\ \hline 

\textsc{after w/ News} &$-$ &$\mathbf{93.9}\pm 0.3$ & $\underline{87.3} \pm 0.7$ /  $\underline{91.0} \pm 0.5$ & $63.9 \pm 2.3$ \\ 

\textsc{after w/ Reviews} &$-$ &$\underline{93.5} \pm 0.3$ & $\underline{86.9}\pm 0.6$ /  $\underline{90.5} \pm 0.5$ &$\underline{65.1} \pm 2.8$  \\ 

\textsc{after w/ Legal} &$-$ &$\underline{93.6}\pm 0.5$ &$\mathbf{87.5}\pm 1.6$ / $\mathbf{90.9} \pm 1.2$  &$\underline{64.8} \pm 1.6$  \\ 

\textsc{after w/ Medical} &$-$ &$\underline{93.3} \pm 0.5$ & $\underline{87.0} \pm 1.1$ /  $90.5\pm 0.7$ & $64.5 \pm 2.1$\\ 

\textsc{after w/ Math} &$-$ &$\mathbf{93.9}\pm 0.4 $ & $\underline{87.3}\pm 1.2$ / $\underline{90.8} \pm 0.9$ & $\mathbf{66.1} \pm 1.9$ \\ 
\Xhline{2\arrayrulewidth} 
\end{tabular}
\caption{Comparison of standard of fine-tuning (\textsc{sft}) and \textsc{after} for \textsc{BERT} (Top) and \textsc{XLNet} (Bottom). 
\underline{Underlined} scores outperform the baseline.
Best scores for each pretrained LM are shown in \textbf{bold}.
We report the mean and standard deviation across five runs on the validation set.}
\label{table:results}
\end{table*}
\endgroup

\section{Results}\label{sec:results}
Table~\ref{table:results} shows the results on the validation sets of the four GLUE datasets for the two pretrained LMs. We compare the two baselines with \textsc{after} using the \texttt{Auxiliary} data from Table~\ref{table:datasets}. We do not report results on \textsc{CoLA} with \textsc{XLNet} ($-$) because the model demonstrated degenerate performance with the available resources for the batch size (see Appendix~\ref{sec:hyperparameters} for more details).

\vspace{2pt}
\noindent\textbf{\textbf{\textsc{BERT}}}.
We observe that the proposed 
approach
(\textsc{after}) outperforms the first baseline (\textsc{BERT} \textsc{sft}) in all four tasks. For most of these tasks, \textsc{after} results in improved performance with every \texttt{Auxiliary} dataset, demonstrating the robustness of our approach across domains. 

Specifically, in \textsc{CoLA}, we observe that fine-tuning with the adversarial loss substantially outperforms standard fine-tuning. 
Specifically, using an \texttt{Auxiliary} dataset from the \textsc{News} domain improves the baseline by $1.8$ points.
In 
\textsc{SST-2},
we notice that 
although standard fine-tuning achieves high accuracy, the use of \textsc{after} still results in slight performance gains ($\! \sim\!0.4 \%$). Similar to \textsc{CoLA}, these improvements are consistent across \texttt{Auxiliary} datasets and often come with reduced variance, compared to \textsc{sft}. 
In \textsc{MRPC}, we observe gains of $1.5$ points on average in accuracy and $1.0$ in F1 over \textsc{sft}. Using \textsc{News} data as \texttt{Auxiliary}, \textsc{after} outperforms the baseline by $2.1$ points in accuracy and $1.5$ in F1.
In  RTE, the proposed approach improves upon the baseline from $64.3 \%$ to $64.8 \%$ in accuracy, using data from the \textsc{Legal} domain. However, we also observe deteriorated performance with the use of some \texttt{Auxiliary} datasets (e.g. \textsc{Medical}, \textsc{Math}). We attribute this result to the similarity between the domain of RTE (Wikipedia) and the domain of the pretraining corpus of \textsc{BERT} (Wikipedia and Books). 
We test this hypothesis in section~\ref{sec:analysis}.

\vspace{2pt}
\noindent\textbf{\textbf{\textsc{XLNet}}}.
We observe in Table~\ref{table:results} that \textsc{after} consistently outperforms standard fine-tuning for an even higher-performing LM (\textsc{XLNet sft}). 

Specifically, in \textsc{SST-2}  \textsc{after} improves the accuracy of standard fine-tuning (\textsc{sft}) by $0.6 \%$ on average and reduces variance, as well. For instance, with the use of \texttt{Auxiliary} data from \textsc{News} or \textsc{Math} domains, \textsc{after} results in $0.9 \%$ improvement in accuracy. In \textsc{MRPC}, the performance boost is also consistent across \texttt{Auxiliary} data. In particular, the use of \textsc{Legal} data leads in absolute improvement of $1.1 \%$ in accuracy and $0.8 \%$ in F1. In \textsc{RTE}, adversarial fine-tuning outperforms the baseline by $1.4 \%$ in accuracy. However, similar to \textsc{BERT}, we observe lower performance when using \textsc{after} with some \texttt{Auxiliary} data (e.g. \textsc{News}, \textsc{Medical}). We attribute this performance degradation to the same reason as \textsc{BERT}, the similarity between the pretraining corpus domain and the target task domain (both LMs have similar pretraining corpora). 

\vspace{2pt}
\noindent\textbf{\textbf{Summary}}. The experiments of this section reveal that \textsc{after} can boost target task performance and reduce variance compared to standard fine-tuning across different pretrained LMs. 
We can therefore attribute the effectiveness of \textsc{after} to 
regularization itself and not to the model architecture. 
We can also observe in Table~\ref{table:results} that the target task performance of our approach (\textsc{BERT after}) is on par (\textsc{RTE}) or higher (\textsc{MRPC}) than using standard fine-tuning with a higher-performing pretrained LM (\textsc{XLNet sft}).
This finding demonstrates the effectiveness of the proposed
approach 
and motivates the need for more effective fine-tuning schemes as a way to improve the target of pretrained LMs on downstream tasks.
\section{Ablation Study}
\label{sec:analysis}
We investigate the effect of some key factors of \textsc{after} such as the relation of the target task domain and the domain of the pretraining corpus of the LM,  the selection of \texttt{Auxiliary} data and the emergence of domain-invariant characteristics. For the experiments of this section we used \textsc{BERT}, unless otherwise stated.

\noindent\textbf{LM pretraining and Task Domains}.
To explore why \textsc{after} fails to improve upon 
the baseline
on \textsc{RTE}, we examine if the pretrained representations are already well suited for the task (i.e.\ no regularization is needed).
We 
calculate the average masked LM (MLM) loss
of \textsc{BERT} for each \texttt{Main} dataset.
We observe in Table~\ref{table:mlm_results} that 
\textsc{SST-2} produces the largest loss
which can be partially attributed to the dataset format (it contains short sentences that make the MLM task very challenging). 
\textsc{RTE} produces the lowest loss 
confirming our intuition regarding the similarity of the pretraining corpus of \textsc{BERT} and \textsc{RTE}. In this case, general-domain 
and domain-specific representations 
are close, rendering domain-adversarial regularization undesirable. This is also confirmed by the the vocabulary overlap between RTE and a Wikipedia corpus (Table~\ref{table:mlm_results}).
The more distant the pretraining domain of \textsc{BERT} is to the specific task (measured by vocabulary overlap and MLM loss), the more benefits \textsc{after} demonstrates, confirming our intuition regarding domain-adversarial regularization.
\begingroup
\setlength{\tabcolsep}{6pt} 
\renewcommand{\arraystretch}{1} 
\begin{table}[h]
\resizebox{\columnwidth}{!}{%
\centering
\small
\begin{tabular}{lcccc}
\Xhline{2\arrayrulewidth}
 &\textbf{RTE} &\textbf{MRPC} &\textbf{CoLA}   &\textbf{SST-2}  \\ \hline

MLM Loss &$\mathbf{2.17}$ &$2.37$ &$2.53$   &$3.39$  \\ 
Overlap with \textsc{Wiki} (\%) &$\mathbf{38.3}$ &$34.0$ &$24.0$   &$26.1$ \\
\textsc{after} Improvement (\%) &$\mathbf{0.8}$ &$2.5$ &$3.2$  &$0.7$  \\
\Xhline{2\arrayrulewidth} 
\end{tabular}}
\caption{Masked LM loss of \textsc{BERT} (the lower the better),
vocabulary overlap with the Wikipedia domain (\textsc{Wiki}) and  improvement of \textsc{after} (best) for each task.}
\label{table:mlm_results}
\end{table}
\endgroup

\begin{figure}[h]
    \centering
  \includegraphics[width=0.58\linewidth]{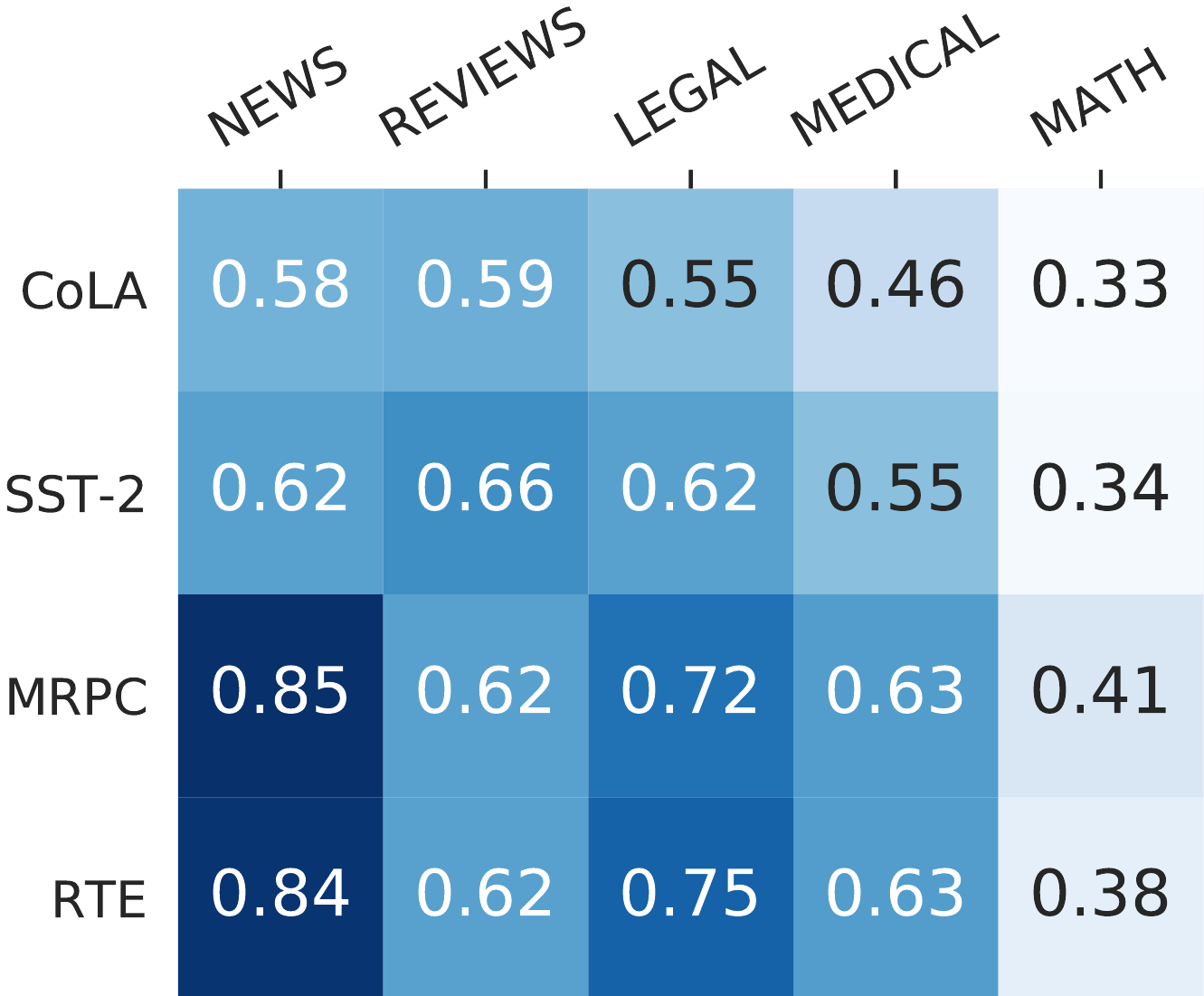}
  \caption{JS divergence between all dataset pairs.}
  \label{fig:js distance}
\end{figure}

\vspace{1pt}
\noindent\textbf{Domain Distance}.
We measure the
domain distance for all \texttt{Main}-\texttt{Auxiliary} pairs 
to evaluate how the choice of 
the latter
affects the performance of \textsc{after}. 
We represent the word distribution of each
dataset using term distributions 
$t \in \mathcal{R}^{|V|}$ where $t_i$ is the probability
of the $i$-th word in the joint vocabulary $V$ 
(see Appendix~\ref{sec:domain dis})
and calculate
Jensen-Shannon (JS) divergence~\cite{plank-van-noord-2011-effective}. 
Combining the results of 
Table~\ref{table:results} and Fig.~\ref{fig:js distance}, no clear pattern emerges demonstrating, perhaps, our method's robustness 
to domain distance. 
We leave a further investigation of selection criteria for the \texttt{Auxiliary} data for future work.

\begin{figure}[t]
    \resizebox{\columnwidth}{!}{%
         \includegraphics[scale=1]{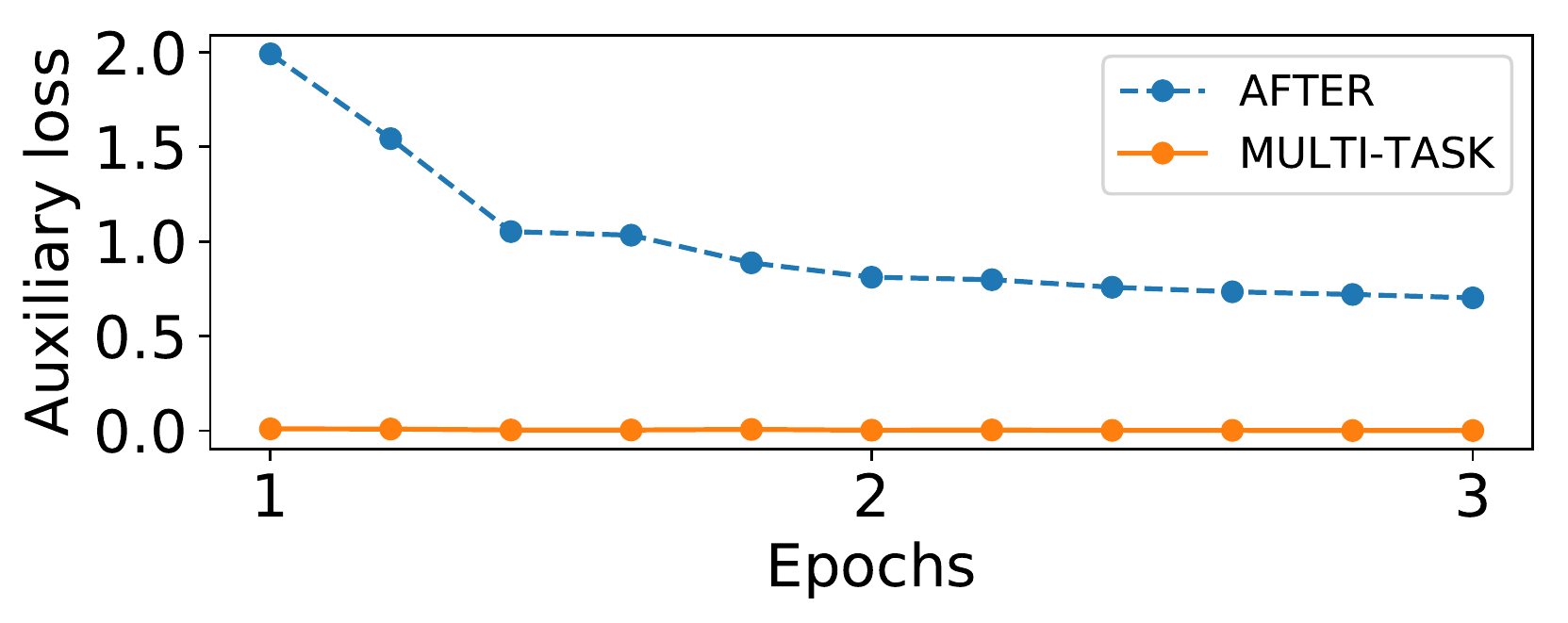}}
        \caption{\textsc{after} ($\lambda>0$) vs.\ \textsc{multi-task} ($\lambda<0$).}
        \label{fig:losses}
\end{figure}

\vspace{1pt}
\noindent\textbf{Domain-invariant vs.\ Domain-specific  Features}.
To investigate if the benefits of \textsc{after} can be attributed only to
data augmentation 
we compare adversarial ($\lambda\!>\!0$ in Eq.~\ref{eq:loss}) and multi-task ($\lambda\!<\!0$) fine-tuning.
We 
experiment with \textsc{MRPC} and 
\textsc{CoLA} for both settings
(tuning each $\lambda$ separately).
We observe that during 
multi-task 
fine-tuning (Fig.~\ref{fig:losses}), $L_{Domain}$ is close to zero 
(even in the first epoch). 
This implies that domain classification is an easy auxiliary task,  confirming our intuition that a non-adversarial fine-tuning setting favors domain-specific features.
Although the multi-task approach leverages the same unlabeled data, its performance is worse than \textsc{after} (Table~\ref{table:multi-task vs after}), which highlights the 
need for an adversarial domain discriminator.
\begin{table}[ht]
\setlength{\tabcolsep}{6pt} 
\renewcommand{\arraystretch}{1} 
\centering
\small
\begin{tabular}{lcc}
\Xhline{2\arrayrulewidth}
 &\textbf{CoLA} &\textbf{MRPC}  \\ \hline 

\textsc{after w/ News} &$\mathbf{57.3}$ &$\mathbf{87.5} / \mathbf{91.1}$ \\ 
\textsc{multi-task w/ News} &$56.5$ &$86.7 / 90.5$  \\ 
\Xhline{2\arrayrulewidth} 
\end{tabular}
\vspace{-1mm}
\caption{Comparison of \textsc{after} vs. \textsc{multi-task}.}
\label{table:multi-task vs after}
\end{table}
\section{Conclusions and Future Work}
We propose \textsc{after}, a domain adversarial method 
 to regularize the fine-tuning process of a pretrained LM.
Empirical results demonstrate that our method can lead to improved performance over standard fine-tuning.
\textsc{after} can be widely applied
to any transfer learning setting and model architecture, with minimal changes to the fine-tuning process, without requiring any additional labeled data.
We aim to further explore the effect of \texttt{Auxiliary} data on the final performance and the use of multiple \texttt{Auxiliary} datasets.
We also aim to extend the proposed approach 
as a way to fine-tune a pretrained LM to a different language, in order to produce language-invariant representations.

\section*{Acknowledgements}
We thank Andrei Popescu-Belis for his valuable comments
and help with an initial draft of this paper.



\bibliography{emnlp2020}
\bibliographystyle{acl_natbib}

\appendix
\clearpage
\appendix

\section{Appendices}
\label{sec:appendix}
In  this  supplementary  material,  we  provide additional information for producing the results in the paper, and results that could not fit into the main body of the paper.

\subsection{Dataset Details} \label{sec:dataset_details}
\noindent\textbf{\texttt{Main} datasets}. 
We use only four datasets of the GLUE benchmark as \texttt{Main} for our experiments, due to resources constraints. All \texttt{Main} datasets are open source and can be found in \url{https://gluebenchmark.com/tasks}.

\noindent\textbf{\texttt{Auxiliary} datasets}.
We choose \texttt{Auxiliary} datasets that are larger than \texttt{Main}, which we consider as the most realistic scenario, given the availability of unlabeled compared to labeled data. We under-sample the \texttt{Auxiliary} dataset  to ensure that the two domains are equally represented, motivated by the observation of ~\citet{bingel-sogaard-2017-identifying}
that balanced datasets tend to be better in auxiliary tasks. For each mini-batch, we sample equally from the \texttt{Main} and \texttt{Auxiliary} datasets. 

The \texttt{Auxiliary} datasets are a mixed of labeled and unlabeled datasets from different domains. The labeled \texttt{Auxiliary} datasets (e.g. \textsc{ag news}) are handled as unabeled corpora, by dropping the task-specific labels and using only the domain labels. Although some domains might seem similar to those of the \texttt{Main} datasets, e.g Electronics Reviews vs. Movies revies and Agricultural News vs. News this is not the case as can be seen in Figure~\ref{fig:tasks_dom_overlap}.

The maximum sequence length for all datasets was 128, so all samples were truncated to 128 tokens and lower-cased.  For \textsc{europarl}, which contains parallel corpora in multiple languages, only the English part is used. We therefore sample 120K sentences from the English corpus. For \textsc{pubmed} we use 120K abstracts from medical papers, from the dataset of \citet{cohan-etal-2018-discourse}. For \textsc{math} we use 120K questions of medium difficulty from  the dataset of \citet{Saxton2019AnalysingMR}. We note that all corpora used are in English.

\subsection{Hyperparameters and Model details} \label{sec:hyperparameters}
For \textsc{BERT} we use the \texttt{bert-base-uncased} pretrained model and we fine-tune it with the following hyperparameters: dropout $0.1$, batch size $28$ and a maximum length of $128$ tokens. For the optimization we use the Adam optimizer ~\cite{DBLP:journals/corr/KingmaB14} with a learning rate of 2e-5, adam epsilon 1e-6 and weight decay $0.01$. We use a linear warmup schedule with $0.1$ warmup proportion. 

For \textsc{XLNet} we use the \texttt{xlnet-base-cased}. We use the last hidden state output embedding, as the input sequence representation. We fine-tune \textsc{XLNet} with the following hyperparameters:  $26$ batch size and the same learning rate (2e-5) and sequence length ($128$) as \textsc{BERT}. We do not use weight decay or warmup. In order to replicate the results of \citet{NIPS2019_8812} in \textsc{CoLA}, the authors suggested using a considerably larger batch size ($\times4$), which was not possible in our case, due to resources constraints\footnote{The authors' response regarding the hyperaparameters : \href{https://github.com/zihangdai/xlnet/issues/96}{Clarification on reported dev numbers on GLUE tasks}. Similar problems have been reported for other \textsc{BERT}-based models on \textsc{CoLA}, as well:  \href{https://github.com/huggingface/transformers/issues/2599}{Xlnet, Alberta, Roberta are not finetuned for CoLA task}.}. 

When we combine \textsc{after} with either \textsc{BERT} or \textsc{XLNet} we use the same hyperparameters as above. We note that both models have approximately $110M$ parameters and this is (almost) the same using \textsc{after}, as well. Our approach only introduces a binary domain discriminator in the form of a linear layer.

For all experiments we used a 6G GeForce GTX 1080. The duration of the experiments depended on the datasets. For  \textsc{SST-2}, which is the largest dataset, the experiments for the baseline (\textsc{BERT}, \textsc{XLNet}) had a runtime of approximately 100mins (for all 4 epochs) and  200mins for \textsc{after}, due to the implicit dataset augmentation. Smaller datasets such as \textsc{MRPC} and \textsc{CoLA} had an approximate runtime of 30mins with standard fine-tuning and 60mins with \textsc{after}.

\subsection{Tuning the $\lambda$ hyperparameter} \label{sec:tuning_lambda}
\begin{figure}[h]
\includegraphics[width=\linewidth]{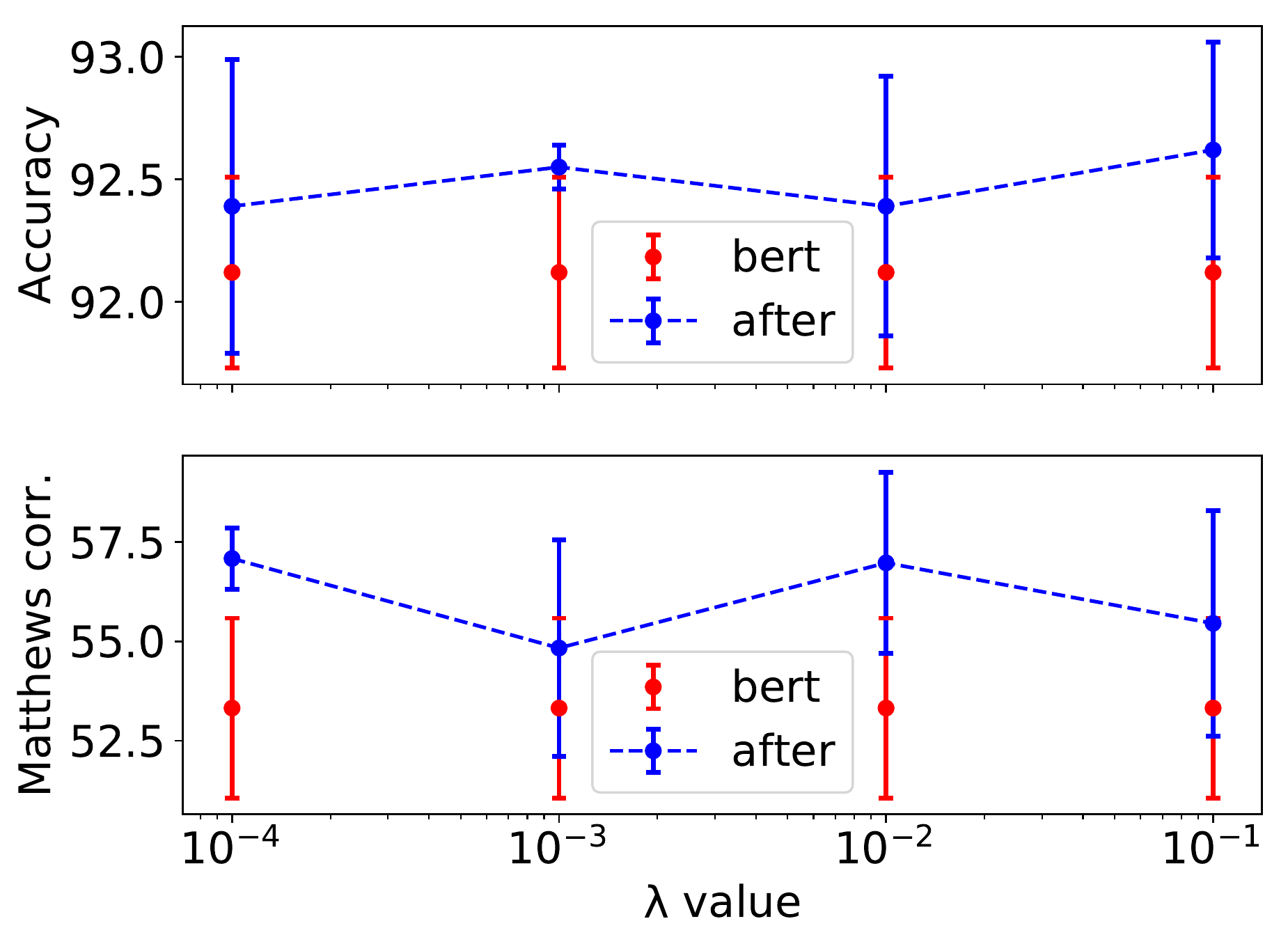}
        \caption{Performance of standard \textsc{BERT} fine-tuning vs. \textsc{after} for different $\lambda$ values on \textsc{SST-2} (Top) and \textsc{CoLA} (Bottom). The errorbars correspond to one standard deviation.}
        \label{fig:lambda_tuning}
\end{figure}

We tune $\lambda$ on each development set, choosing from  $\{0.1, 0.01, 0.001, 0.0001\}$. In Figure \ref{fig:lambda_tuning} we compare the performance of \textsc{BERT} and \textsc{after} for different \texttt{Main}-\texttt{Auxiliary} combinations, as we vary the value of $\lambda$.

We observe that the various values of $\lambda$ can have different effect on the performance and variance of \textsc{after}. We observe that most values of $\lambda$ significantly improve the performance of the baseline, \textsc{BERT} and an exhaustive search is not required. 
Table \ref{table:lambda_results} presents the values of $\lambda$ that were used for the results reported in Table \ref{table:results}. Best values of $\lambda$ were chosen based on the task-specific metric (e.g. Accuracy, Matthews correlation).
\begingroup
\setlength{\tabcolsep}{6pt} 
\renewcommand{\arraystretch}{1} 
\begin{table*}[h]
\centering
    \begin{tabular}[t]{|l|l|l|l|l|}
    \Xhline{2\arrayrulewidth}
    \multicolumn{5}{|c|}{\textsc{BERT}} \\ \hline
         
     &\textbf{CoLA} &\textbf{SST-2} &\textbf{MRPC} &\textbf{RTE} \\ \hline
    
    \textsc{News} &0.1 &0.1 &0.1 &0.001 \\ 
    
    \textsc{Reviews} &0.1 &0.01 &0.1 &0.001 \\ 
    
    \textsc{Legal} &0.01 &0.1 &0.01 &0.1  \\ 
    
    \textsc{Medical} &0.01 &0.1 &0.1 &0.0001 \\ 
    
    \textsc{Math} &0.001 &0.1 &0.001 &0.001 \\ 
    \Xhline{2\arrayrulewidth} 
    \end{tabular}
    \begin{tabular}[t]{|l|l|l|}
    \Xhline{2\arrayrulewidth}
    
    \multicolumn{3}{|c|}{\textsc{XLNet}} \\ \hline
    \textbf{SST-2} &\textbf{MRPC} &\textbf{RTE} \\ \hline
    
    0.01 &0.1 &0.0001 \\ 
    
    0.001 &0.0001 &0.01 \\ 
    
    0.001 &0.01 &0.0001  \\ 
    
    0.1 &0.1 &0.01 \\ 
    
    0.0001 &0.01 &0.01 \\
    
    \Xhline{2\arrayrulewidth} 
    \end{tabular}
\caption{Best $\lambda$ value of \textsc{after} for each experiment.}
\label{table:lambda_results}
\end{table*}
\endgroup

\subsection{More Domain Distance Results} \label{sec:domain dis}
In order to create a common vocabulary for all data for Figure ~\ref{fig:js distance} we find the $5k$ most frequent words in each dataset and we then take the union of these sub-vocabularies which results in $23k$ words. We also calculate the vocabulary overlap, by creating each domain (or task) vocabulary with  the $10k$ most frequent words in each dataset (in case a dataset contains less words we use all the words in the dataset). 

\begin{figure}[h]
  \includegraphics[width=\linewidth]{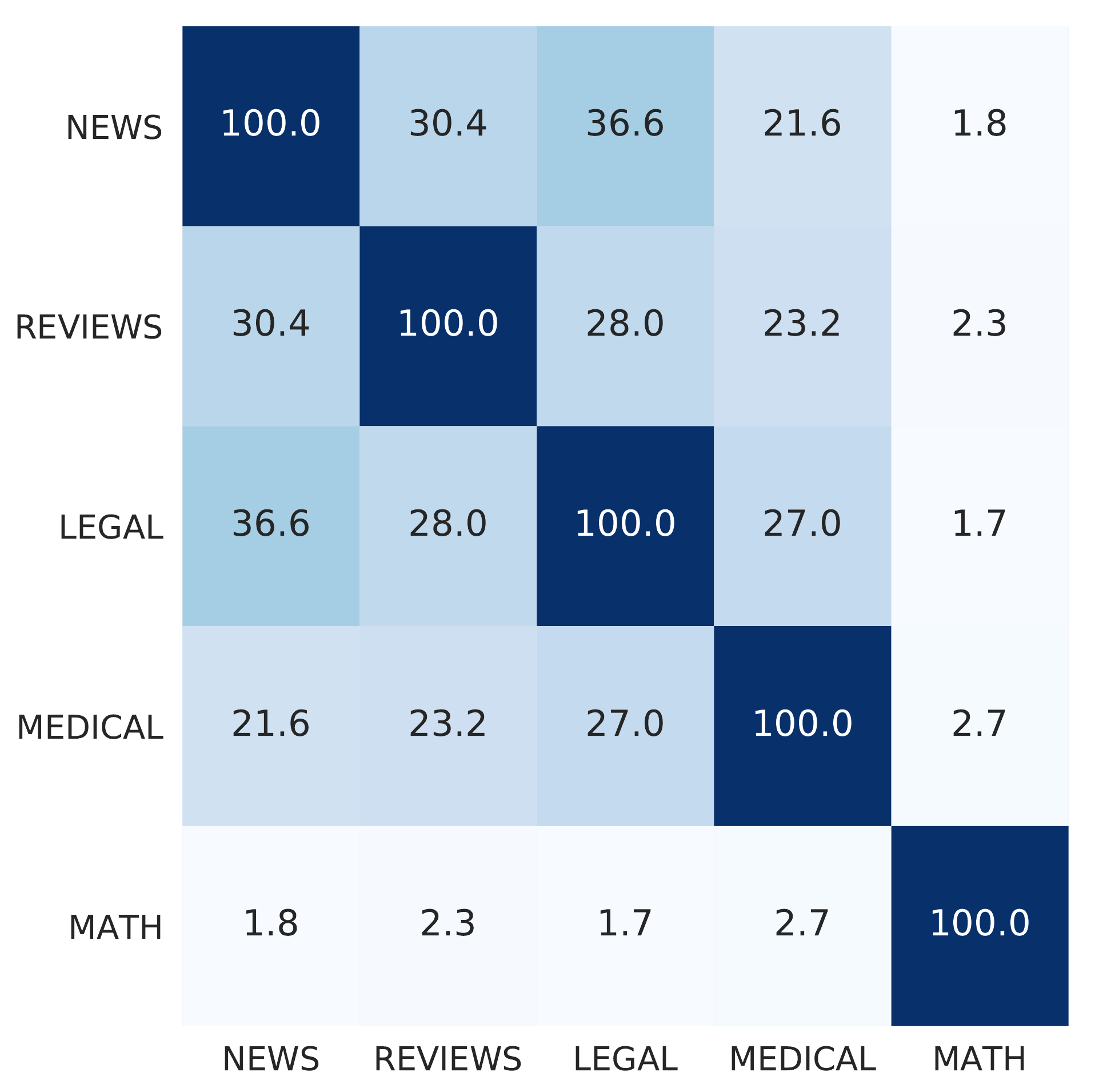}
  \caption{Vocabulary overlap (\%) between domains.}
  \label{fig:vocab_overlap}
\end{figure}

\begin{figure}[h]
  \includegraphics[height=4.92cm, keepaspectratio]{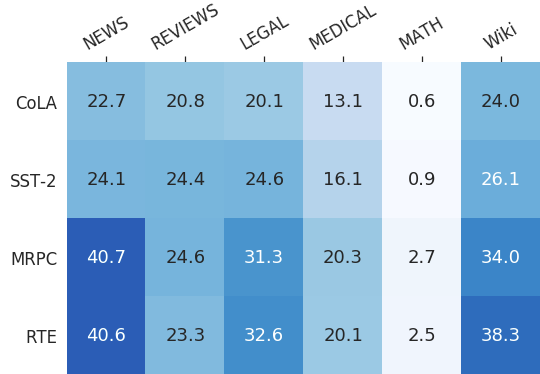}
  \caption{Vocabulary overlap (\%) between tasks and domains.}
  \label{fig:tasks_dom_overlap}
\end{figure}

We then calculate the vocabulary overlap between domains (Figure~\ref{fig:vocab_overlap}) and between each task and all domains (Figure~\ref{fig:tasks_dom_overlap}). For the latter, we also include the \textsc{Wiki} domain to account for the pretraining domain of \textsc{BERT} and \textsc{XLNet}. For the vocabulary of \textsc{Wiki} we use the WikiText-2 corpus from \citet{DBLP:conf/iclr/MerityX0S17}. We observe in Figure~\ref{fig:vocab_overlap} that most domains are dissimilar, with the exception of \textsc{news} and \textsc{legal} domains, that have $36.6 \%$ vocabulary overlap. In Figure~\ref{fig:tasks_dom_overlap}, we observe that \textsc{rte} has the most overlap in vocabulary with \textsc{Wiki} which is a possible cause for the deteriorated performance of \textsc{after}, since the model has already been pretrained in this domain and does not require further regularization, as described in Section~\ref{sec:analysis}. 

\end{document}